\documentclass{article}

\usepackage{microtype}
\usepackage{graphicx}
\usepackage{booktabs} % for professional tables
\usepackage[dvipsnames]{xcolor}
\usepackage[acronym]{glossaries}
\usepackage{graphicx}
\usepackage{array}
\usepackage{booktabs}
\usepackage{amssymb}
\usepackage{caption}
\usepackage{makecell}
\usepackage{hyperref}

\usepackage[accepted]{styles/icml2026}
\makeatletter
\renewcommand{\ICML@appearing}{}
\makeatother

\usepackage{amsmath}
\usepackage{amssymb}
\usepackage{mathtools}
\usepackage{amsthm}

\usepackage[capitalize,noabbrev]{cleveref}

\theoremstyle{plain}

\theoremstyle{definition}

\theoremstyle{remark}

\usepackage[textsize=tiny]{todonotes}

\newacronym{senn}{SENN}{Self-Explaining Neural Network}
\newacronym{ai}{AI}{Artificial Intelligence}
\newacronym{xai}{XAI}{Explainable Artificial Intelligence}
\newacronym{xrl}{XRL}{Explainable Reinforcement Learning}
\newacronym{cps}{CPS}{Cyber-Physical Systems}
\newacronym{qoe}{QoE}{Quality of Experience}
\newacronym{dl}{DL}{Deep Learning}
\newacronym{dnn}{DNN}{Deep Neural Network}
\newacronym{ppo}{PPO}{Proximal Policy Optimization}
\newacronym{iml}{IML}{Interpretable Machine Learning}
\newacronym{rl}{RL}{Reinforcement Learning}
\newacronym{ue}{UE}{User Equipment}
\newacronym{bs}{BS}{Base Station}
\newacronym{drl}{DRL}{Deep Reinforcement Learning}
\newacronym{ed}{ED}{Effects Distributions}

\icmltitlerunning{Submission for AI4NextG 2026}

\begin{document}

\twocolumn[
  \icmltitle{Self-Explaining Reinforcement Learning for \\ Mobile Network Resource Allocation}



  \begin{icmlauthorlist}
    \icmlauthor{Konrad Nowosadko}{ericsson}
    \icmlauthor{Franco Ruggeri}{ericsson}
    \icmlauthor{Ahmad Terra}{ericsson}
  \end{icmlauthorlist}

  \icmlaffiliation{ericsson}{Ericsson Research}

  \icmlcorrespondingauthor{Konrad Nowosadko}{konrad.nowosadko@ericsson.com}
  \icmlkeywords{Machine Learning, ICML}

  \vskip 0.3in
]

\printAffiliationsAndNotice{}

\begin{abstract}
Deep reinforcement learning (DRL) methods, though powerful, often lack transparency, which limits their adoption in critical domains. We apply Self-Explaining Neural Networks (SENNs) to RL by parametrizing the policy of a PPO agent with a SENN, producing intrinsic local explanations, and propose a method for aggregating them into global explanations. We evaluate our approach on a mobile network resource allocation problem, our approach performs within a small margin of the state-of-the-art deep learning method and significantly outperforms the best deployed heuristic, while the extracted global explanations correlate strongly with DeepLift and InputXGradient, making SENNs a promising candidate for high-stakes RL.

\end{abstract}

\section{Introduction}

%MOVE1: XAI 
Although powerful, \gls{ai} models often operate as black-boxes, making it difficult to interpret their decisions, leading to a lack of trust among stakeholders and consequently hindering their applicability. This lack of transparency of black-box models, such as \glspl{dnn}, limits their applicability in high-stakes domains \cite{linardatos_explainable_2021}. This limited applicability facilitates rapid growth in the interest of \gls{xai}. \gls{xai} is divided into  post-hoc interpretability methods and intrinsic interpretability methods. 
Post-hoc interpretability methods attempt to explain trained black-box models, by analysing relations between model's inputs and outputs. For instance, LIME or SHAP, are post-hoc perturbation-based methods that compute feature attributions representing the impact of each feature on the output of the model \cite{conceptual_challenges, GradSHAP, LIME}. On the other hand, intrinsic interpretability means that the \gls{ai} method is interpretable by design, for instance, linear regression, where the contribution of each feature towards the final decision is expressed by the corresponding coefficient.   Moreover, intrinsic interpretability methods are considered more robust as they provide explanations derived from the model's internal mechanism \cite{Rudin}. \gls{xai} methods can be further divided into local and global methods. Local providing explanations to  individual predictions and global that provide explanations of the whole model. Extending the prior example, linear regression models provide global intrinsic explanations.

The intuitions that ground \gls{xai} also extend to \gls{rl} through \gls{xrl}. To date, many \gls{xrl} specific techniques have been proposed, including temporal policy decomposition and hierarchical skill acquisition \cite{temporal_policy_decomp,hierachical_skill}. However, most state-of-the-art \gls{rl} algorithms are model-free and rely on \glspl{dnn} for policy learning and modelling; because \glspl{dnn} lack transparency, these algorithms inherit the same opacity \cite{xrl_survey_2}.

This work focuses on improving the explainability of the used models within the \gls{rl} formulation, improving the explainability of the whole system and introduces a new approach for \gls{xrl}. We apply \glspl{senn} to \gls{rl} networks that match \gls{dnn} expressiveness, while offering adjustable interpretability \cite{senns}. The main contributions of the paper are following:
\begin{itemize}
    \item Applying and evaluating \glspl{senn} on \gls{rl} problem and telecom use case. 
    \item Modifying the architecture of \glspl{senn} to improve explainability and robustness. 
    \item Proposing a method to extract global explanations from the local \gls{senn} explanations.
\end{itemize}

\section{Related work}

% State-of-the-art RL algorithms (e.g., PPO) are based on function approximation with DNNs --> lack of interpretability
\glspl{senn} are intrinsically interpretable neural networks that decompose their output as a linear combination of learned input concepts weighted by input-dependent relevance scores \cite{senns}. The most recent work on \glspl{senn} has extended the original formulation in several directions: C-SENN \cite{csenn} introduced contrastive training to improve concept disentanglement, while Q-SENN \cite{qsenn} applied quantization to enforce sparse, binary concept-feature relationships that improved both accuracy and human interpretability on complex vision tasks. These studies did not attempt to extract global explanations from \glspl{senn} or or apply them to \gls{rl} setting. 

Existing \gls{xrl} methods approach explainability from two angles \cite{milani_survey}. The first angle focuses on explaining RL components such as states, rewards, or trajectories. Some of the well-known methods within this approach include reward decomposition and Shapley values calculated on states \cite{beechey_explaining_2023,reward_decomposition,ruggeri_rollout-based_2024}. The second angle instead addresses the dominant source of opacity in state-of-the-art \gls{rl} algorithms that stems the deep neural networks — such as those used in DQN or PPO — that parametrize policy or value functions \cite{xrl_survey_2,PPO, DQNs}.Within this approach the main line of research use more explainable models or explain model's decisions through popular model agnostic methods such as SHAP, DeepLift or InputxGradient, which estimate input feature attribution towards model's final decision \cite{DeepLiftIXG,inputxgradient,GradSHAP, exploratory,lmut,electrical_tilt}. In this work, we follow the line of research on intrinsic explainability by applying SENNs to RL policies. Unlike previous work, which focuses on replacing deep models with less expressive tree-based and linear models, SENNs have the potential to be as performant as ordinary deep models while bringing enhanced explainability.

Proposed methods are evaluated using \textit{mobile-env} a simulation environment introduced in \cite{mobile-env} that simulates a resource allocation problem in mobile networks. Currently the optimal solution for mobile-env is based on PPO algorithm where actor and critic are parametrized using \glspl{dnn}. \gls{rl}, despite its proven superior performance, to heuristics-based solutions, is not yet widely used in the Coordinated Multipoint problem and, to the best of our knowledge most infrastructure still relies on heuristics \cite{schneider2023deepcomp}. Most likely due to their reliability and transparency, despite their poor scaling and poorer performance in longer time-horizons \cite{schneider2023deepcomp}.

\section{Self-explaining reinforcement learning}
\glspl{senn} consist of three modules: conceptizer, parametrizer, and aggregator. The conceptizer extracts features, providing the concept vector $\mathbf{h}$, the parametrizer  produces a relevance score vector $\mathbf{R}$. Lastly, the aggregator aggregates the concept vector and relevance score matrix, and is implemented through dot product. Final output, is expressed as:
\begin{align}
\label{reg_senns}
  %sf(x) = g(\theta(x), h(x)) = \theta(x)^Th(x) = \mathbf{R}_{C\times m} \cdot \mathbf{h}_{m \times 1} = \mathbf{\hat{y}}_{C \times 1}
  f(x)  = \theta(x)^Th(x) = \mathbf{R}_{C\times m} \cdot \mathbf{h}_{m \times 1}
\end{align}
where, $\mathbf{R}$ are relevance scores matrix, $\mathbf{h}$ is concept vector, $C$ and $m$ are number of classes and number of concepts respectively.

The explainability of \glspl{senn} originates from the relevance scores $\mathbf{R}$ and local stability of its relevance scores with regard to the concept vector. Each concept is weighted by the corresponding relevance score, providing us with the attribution of each feature towards the final output.  Local stability forces similar attributions for samples with similar concept values. To measure the stability of explanations, Lipschitz continuity is leveraged with Lipschitz constant, defined as:
%Local stability is evaluated using the Lipschitz constant: 
\begin{equation}
\label{lipshitz}
 L = \max_{x \in X}\frac{\|f(x) - f(x_0)\|}{\|h(x) - h(x_0)\|}, 
\end{equation}
where $f(x)$ is a \gls{senn} model, $h(x)$ is a conceptizer. 

Due to the computational complexity of calculating the exact value of $L$, authors of \citep{senns}, use stochastic gradient descent to estimate it.
To ensure explainability, \glspl{senn} are trained with  a loss function, $\mathcal{L}_y(f(x), y) + \lambda \mathcal{L}_\theta(f) + \xi \mathcal{L}_h(x, \hat{x})$, that allows to influence the local stability of the model through adjustment of $\lambda$. $\mathcal{L}_\theta(f)$ is a robustness loss, $\mathcal{L}_y(f(x), y)$ is a classification loss, the last term is conceptizer's reconstruction loss \cite{senns}. 

% On sentance introducing the Self explaining reinforcement learning
\subsection{Modifications}

In this work, we consider low-dimensional, non-visual domain where the input dimensions are interpretable. Motivated by this setting, we adopt an identity conceptizer that uses the input features directly as concepts for the explanations. %The SENN model we apply has one architectural change compared to its default form \cite{senns}, motivated by our problem setting. Following~\cite{senns}, we adopt an identity conceptizer: in low-dimensional, non-visual domains, the concept explanations from \cite{senns} become less explicit, and feature extraction provides no benefit when the input dimensions are themselves interpretable. 
Beyond this, we introduce a trainable bias vector in the aggregator. We hypothesize that the bias term can absorb the state-independent component of the policy's action preferences, allowing the relevance scores $\theta(s)$ to represent state-specific deviations from the baseline rather than encoding both signals jointly, which could result in noisier explanations.

\begin{figure}[ht]
  \vskip 0.2in
  \begin{center}
    \centerline{\includegraphics[width=\columnwidth]{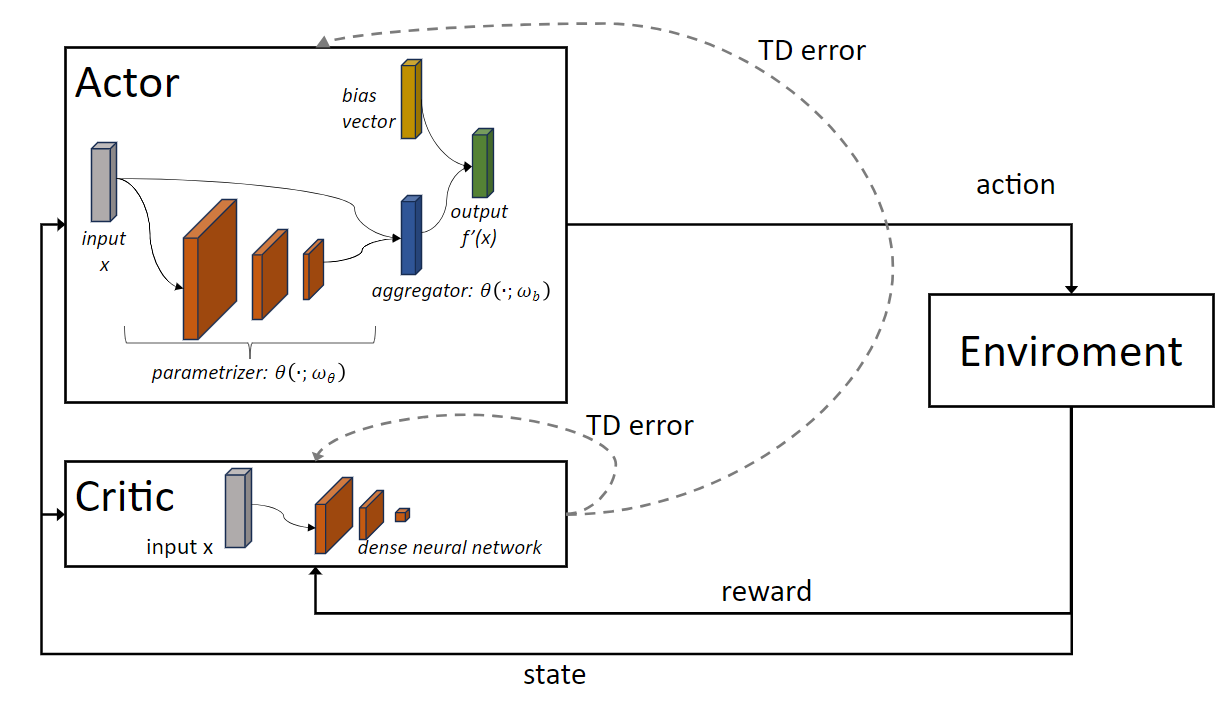}}
    \caption{Overview of the method, the \gls{senn} is an actor and models the policy. Critic, evaluating actor decision is implemented via the \gls{dnn} \cite{PPO}. The environment takes in the actor's action and returns the state and reward.}
    \label{fig:arch_senns}
  \end{center}
\end{figure}

With these modifications we can rewrite Equation \eqref{reg_senns} as follows:
\begin{equation}
\label{mod_senns}
    f'(x) = \theta(x)^Tx+b,
\end{equation}
where $b$ stands for the bias term. We aim to obtain an explainable policy learned by the actor, leaving value prediction uninterpretable as it is not used during inference. Thus, the actor, policy is implemented via \gls{senn} and critic, the value function via \gls{dnn}.
Fig. \ref{fig:arch_senns} presents the \glspl{senn}' architecture and the \gls{ppo}'s diagram. 

\subsection{Local explanations}
\glspl{senn} provides local intrinsic explanations, which are sourced from the parametrizer's relevance score. Relevance scores weigh the input features, providing a numerical measure of the feature's contribution towards the final decision, similar to linear models. 
For each decision, the local explanation consists of the relevance scores $\theta(x)$ and the effect scores. The effect scores are defined as follows, inspired by linear models' explanations \cite{molnar}:
$
\label{effects_def}
E_{i} = \theta(x_{i}) \odot x, \quad \forall i \in \{1, \dots, m\},
$
where $\theta \in \mathcal{R}^{m \times n}$ and $x \in \mathcal{R}^{n \times 1}$, $n$ is input size and $m$ is number of output classes and $\odot$ is the Hadamard product. The effects are relevance scores scaled by the feature values incorporating information of direct feature's contribution. Relevance scores reflect feature importance, while effects capture their direct contribution to the predictions, together they offer a more complete explanation of model's decision. Furthermore, the lack of black-box conceptizer improves the clarity of the explanations, discarding the need for often dubious concept explanations \cite{csenn}.

\subsection{Global explanations}
\label{global_explanations}

While local explanations can clarify individual decisions made by the model, they do not capture the model's general behaviour like global explanations do. We therefore propose a novel clustering-based method for extrapolating global explanations based on the local explanations naturally provided by \glspl{senn}. To create global explanations we first collect set of decisions for which we extract local explanations, then we aggregate those explanations creating more general, global explanations. Then, we calculate \emph{sway scores}, defined as follows:
\begin{equation}
    s_i(x) = \theta_i^{a^*} \cdot h_i(x) \;-\; \max_{a \neq a^*} \bigl(\theta_i^{a} \cdot h_i(x)\bigr)
    \label{sway_eq}
\end{equation}
A sway score expresses a decision margin of the i-th concept i.e., how much concept i-th argues for action $a^*$ in comparison to the highest i-th concept value from the remaining actions. Unlike relevance scores, sway scores aim to capture the impact on the decision itself, relative to other actions, rather than the magnitude of concept activation. Finally, we cluster sway scores via k-means and average within each cluster, this preserves the structure that would be lost when averaging across all samples of an action. Additionally, we can aggregate the clustering results to create attribution scores (see section \ref{exp_evaluation}). This allows us to validate our method by comparing it against state-of-the-art post-hoc techniques.

\section{Experimental setup}

\subsection{Mobile-env}
We apply our method to mobile-env, a simulator that implements a resource allocation problem in mobile networks \cite{mobile-env}. In this environment, base stations (BSs) are distributed over an area and provide mobile connectivity to, randomly moving, user equipment (UE), which rely on the BSs to maintain network access. %The UEs move randomly across the area at a constant speed, during which they will connect to the BSs. For each episode, the positions of the BSs remain fixed, while the initial locations of the UEs are sampled from a uniform distribution over the simulation area. 
The goal is to determine a sequence of BS–UE assignments that maximizes the overall utility, defined as the cumulative \gls{qoe} across all UEs. Finding the exact solution to this problem is a challenge even for the small instance; thus, existing solutions rely on simple heuristics or deep reinforcement learning \cite{multiagent}. 
Following centralized training with decentralized execution, we train a single model on data aggregated from all UEs, then deploy a copy to each UE independently.
% mention maybe that it is multi agent system.
We approach a small instance of the problem with three UEs and three BSs. 
Each UE observes a 13-element vector comprising connection statuses, SNR values, its own utility, and the load and utility of each BS. The reward is \gls{qoe}, a logarithmic score in $(-10, 10)$, and the single action available to each UE is to connect to or disconnect from a chosen BS.
%The input space observed by UEs is thirteen-element vector, whose content is explained in Table \ref{tab:observations}, along with the actions' and rewards' description.

% \begin{table}[t]
%     \centering
%       \caption{Output of the environment, environment outputs both rewards and observations, $n$ is a number of base stations.}
%     \resizebox{\columnwidth}{!}{%q
%         \input{resources/tables/env_output}
%     }

%     \label{tab:observations}
% \end{table}

\subsection{Experiments}
We trained a biased \gls{senn} model, a \gls{senn} model, and a \gls{dnn} model, using the \gls{ppo} algorithm with training of 400,000 timesteps across 2 parallel, randomly seeded simulations each consisting of 3 UEs and 3 BSs we decided on episode length of 1200 timesteps. We trained models using \gls{ppo} and hyperparameters reported by \cite{schneider2023deepcomp} except: entropy coefficient of 0.001 and GAE-$\lambda$ of 0.99. \gls{dnn}-based policy was modeled by 3-layer, 256 hidden units each, neural network, \gls{senn}-based policies had parametrizer network of the same size and identity conceptizer. Value network for all cases was  modeled by 4-layer, 256 hidden units each \gls{dnn}.

\subsection{Evaluation}
\label{exp_evaluation}

To evaluate performance of trained models on the resource allocation task we evaluated it over 18,000 timesteps. This evaluation was run on fifteen differently seeded episodes each of 1200 timesteps. Furthermore, to compare effect of local stability and robustness factor on the predictive performance we ran training with different $\lambda$ values and estimated Lipschitz constant for each model using gradient descent, in accordance with \cite{senns}. Trajectories collected during performance evaluation were first filtered, we removed toggle actions that were not applied to the environment due to \gls{ue} being out of range of the \gls{bs} range, and used them to generate explanations. Local explanations were taken from randomly chosen timestep. To generate \emph{global explanations} we calculated sway scores for each decision in filtered trajectories and clustered sway scores using K-means algorithm on the arbitrarily chosen range of $k=\{2,3,4,5,6\}$. For each action we selected number of clusters with the highest silhouette score. After clustering to have more quantifiable measure of explanations faithfulness we decided to compare explanations with well-established existing post-hoc methods: GradSHAP, DeepLift and InputXgradient \cite{GradSHAP,DeepLiftIXG,inputxgradient}. We generated clustering-based attributions by averaging sway scores over clusters for each action.

\section{Results and discussion}
In this section we present the performance results of \gls{senn} and biased \gls{senn} against baselines, the stability–accuracy trade-off, and exemplary local and global explanations.

\subsection{Performance comparison}

We first benchmark \glspl{senn} against \gls{dnn} and heuristic baselines to confirm predictive performance worth explaining. We trained all deep models with PPO. For a faithful comparison we first tried to match the episodic return reported in \cite{schneider2023deepcomp}, using \glspl{dnn}. Then with the same hyperparameters we trained \glspl{senn} variants. Lastly, we also compared obtained results with currently used and best performing heuristic, Dynamic Selection \cite{schneider2023deepcomp}.

\begin{table}[t]
  \caption{Comparison of mean and median episodic return from 15 episodes each 1200 timesteps.}
  \label{sample-table}
  \begin{center}
    \begin{small}
      \begin{sc}
        \begin{tabular}{lcc}
          \toprule
          Model  & Mean return & Median return \\
          \midrule
          DNN               & 1.803 $\pm$ 0.247 & 1.890\\
          SENN              & 1.805 $\pm$ 0.200 & 1.807\\
          Biased SENN       & 1.780 $\pm$ 0.239 & 1.799\\
          Dynamic Selection & 1.428 $\pm$ 0.219 & 1.456\\
          \bottomrule
        \end{tabular}
      \end{sc}
    \end{small}
  \end{center}
  \vskip -0.1in
\end{table}

Tab. \ref{sample-table} shows results of the performance evaluation over $15$ episodes. \gls{dnn} model is performing the best, all deep models significantly outperform Dynamic Selection($\epsilon=0.5$) and biased \gls{senn} and \gls{senn} have similar performance. Small performance gap between \gls{dnn} and \glspl{senn}, is most likely a result of the stability imposed on the \glspl{senn} through the robustness loss. For this experiment robustness loss factor was set to $\lambda=0.001$. 

\begin{figure}[ht]
  \vskip 0.2in
  \begin{center}
    \centerline{\includegraphics[width=\columnwidth]{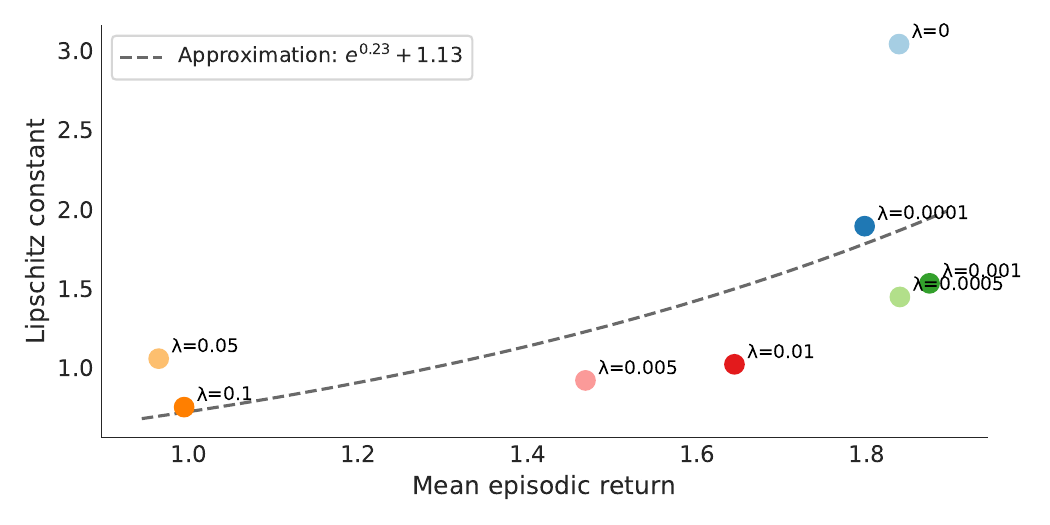}}
    \caption{Relation between, different $\lambda$ values, Lipschitz constant and episodic return (averaged over 3600 timesteps).}
    \label{fig:lipschitz_resource_allocation}
  \end{center}
\end{figure}

Local stability forces \glspl{senn} to learn relevance scores consistent with concept values, but enforcing it introduces the robustness loss, a strong regularizer. Understanding the relation between episodic reward and local stability (estimated via $L$, Eq.~\eqref{lipshitz}) is a key to selecting an appropriate $\lambda$ and assessing the cost of explainability on predictive performance. Fig.~\ref{fig:lipschitz_resource_allocation} presents, the relation between stability and episodic rewards. Each, biased \gls{senn} model was trained with different $\lambda$ and evaluated on $3600$ timesteps, estimation of $L$ is calculated based on the trajectories collected during the evaluation. It shows trade-off between local stability and performance, episodic returns drop with an increasing Lipschitz constant and $\lambda$. The relation appears to be linear in nature; however, its character can be problem dependent as hinted by \cite{senns}.

% Discussion of the section here.

Results, in line with \cite{senns}, confirm that \gls{senn}-based models almost match the performance of similarly sized \glspl{dnn}, and thus \cite{schneider2023deepcomp}, demonstrating successful application to resource allocation in mobile networks and \gls{rl} more broadly. They additionally offer adjustable explainability via local stability and $\lambda$. The bias term, in biased \glspl{senn}, has negligible effect on the performance. Overall, \glspl{senn}-based models seem to be a good alternative to \glspl{dnn}, even though they perform slightly worse than \glspl{dnn}, they offer intrinsic explainability.

\subsection{Local explanations}

\begin{figure}[ht]
  \vskip 0.2in
  \begin{center}
    \centerline{\includegraphics[width=0.7\columnwidth]{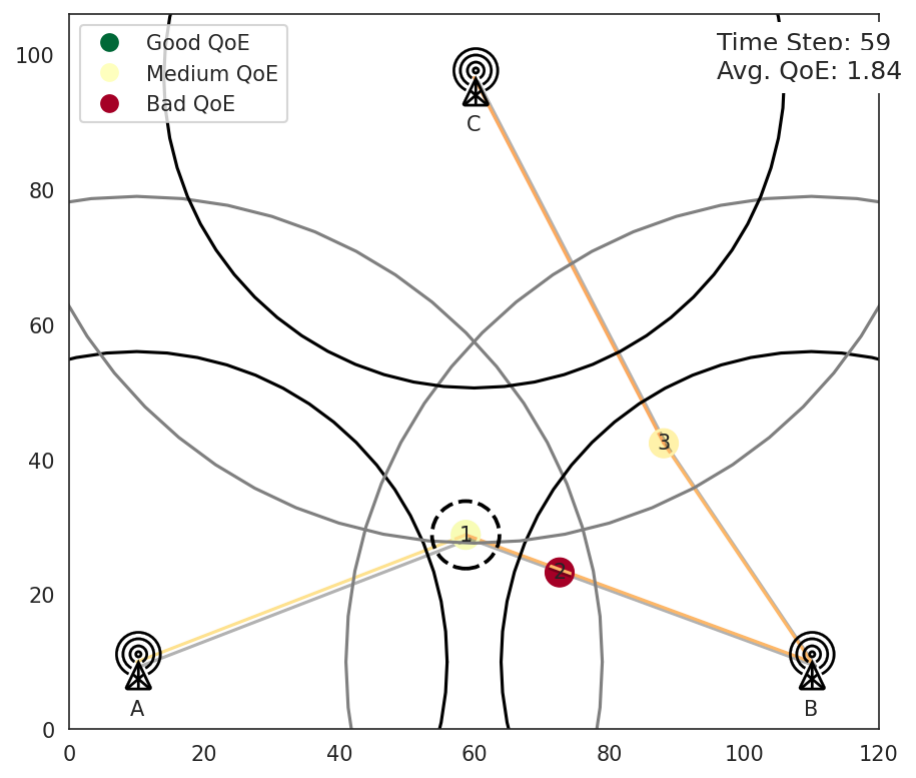}}
    \caption{Visualization of the environment, timestep 59.}
    \label{input_local}
  \end{center}
\end{figure}

\begin{figure}[ht]
  \vskip 0.2in
  \begin{center}
    \centerline{\includegraphics[width=\columnwidth]{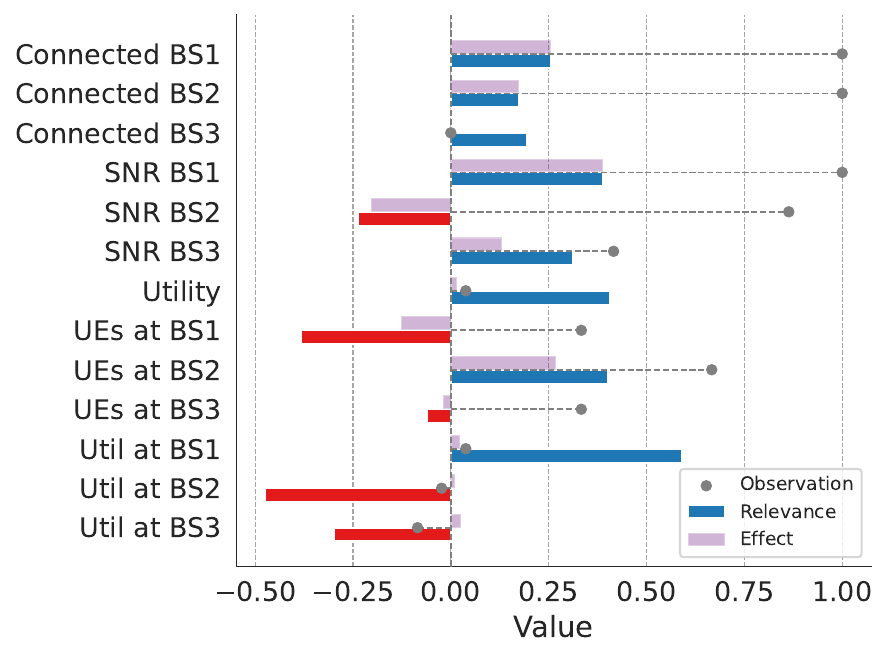}}
    \caption{Relevance scores, effect scores and observations corresponding to the timestep 59, \gls{ue}1 and action toggle \gls{bs}1, see Fig. \ref{input_local}.}
    \label{BS1_local}
  \end{center}
\end{figure}

% 1. Two types of explainability, both stem from the relevance score visualization. The explanation are explaining policy function. 
\glspl{senn}-based models' explainability mainly stems from exposing the relevance scores, a set of input-dependent coefficients that weigh concepts. Visualizing relevance scores shows direct contribution of concepts for each individual decision. 

We present observations for a randomly chosen timestep and \gls{ue} in Fig. \ref{BS1_local} and their visualization in Fig. \ref{input_local}. Chosen \gls{ue} 1 is connected to base \glspl{bs} A (1) and B (2), in that timestep it made a decision to disconnect from \gls{bs} B. Relevance scores and effect scores for that particular decision are presented in Fig. \ref{BS1_local}, effects are concepts scaled by the relevance scores, see Eq. \eqref{effects_def}. Fig. \ref{BS1_local}, shows concepts that contributed strongly positive (connection with \gls{bs}1, connection with \gls{bs}2, SNR of \gls{bs}1, SNR of \gls{bs}3, UE's utility, number of UEs at \gls{bs}2 , Utility at \gls{bs}1) and concepts that contributed negatively (SNR of \gls{bs}2, number of UEs at \gls{bs}1). These relevance scores provide a fairly clear idea of how concepts participated in making a decision at a given timestep. Lastly, contribution of each concept does not seem to be surprising and corresponding relevance scores seem reasonable, \emph{hypothetical} an interpretation of the reason behind this action is that, \gls{ue} disconnected from the BS2 due to stronger signal of \gls{bs}1 and its higher utility as well as low own utility, despite strong signal from \gls{bs}2.  

\glspl{senn}-based models provide understandable, intrinsic explanations that can be accessed for every individual decision. Even though \glspl{senn} are deep models, due to the local stability, relevance scores are forced to behave linearly locally in respect to concepts. In conclusion \glspl{senn} provide intrinsic, per-decision understandable explanations with locally linear relevance scores.% Last sentence can be removed

\subsection{Global explanations}
% 1. We also need global explanations to be able to explain model's general behaviour. 
Local explanations, while highly useful, do not provide insights into the model's general behavior that global explanations offer. In this section, we present results of applying the proposed clustering-based method of aggregating concept sways, see Eq. \eqref{sway_eq}, across multiple runs to explain the general behavior of \glspl{senn}.

\begin{figure}[ht!]
  \vskip 0.2in
  \begin{center}
    \centerline{\includegraphics[width=\columnwidth]{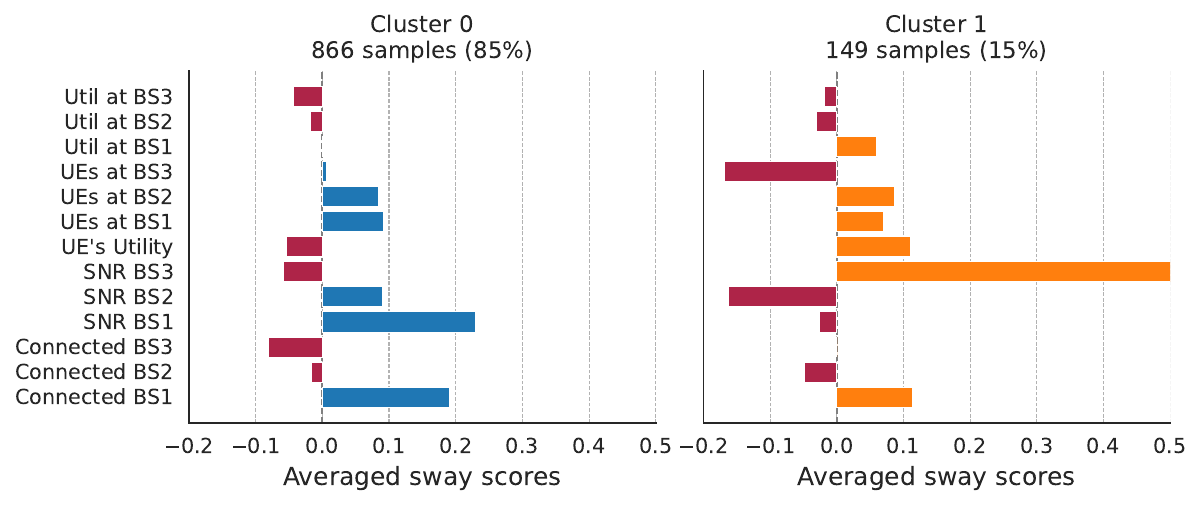}}
    \caption{Mean sway scores per concept for the toggle \gls{bs}3 action, obtained by clustering sways collected over 18\,000 evaluation timesteps and averaging within each cluster.}
    \label{fig:global_average}
  \end{center}
\end{figure}

% 2. Give clustering, why clustering and not for instance averaging. 
We used biased \gls{senn} trained with $\lambda=0.001$. After collecting the data we clustered sway scores by action. We then averaged sway scores within the resulting clusters to obtain typical concept sways for each action. Fig. \ref{fig:global_average} presents two clusters with typical sways for the toggle \gls{bs}3 action. In Cluster 1, the SNR strongly sways the model toward changing the connection status with \gls{bs}2, while the SNR of \gls{bs}2 and the number of \glspl{ue} at \gls{bs}3 sway against the action. In cluster 0, however, the average sway scores are considerably more surprising, with no significant contributions except from the SNR of \gls{bs}1 and the connection status to \gls{bs}1. Given a Silhouette score of 0.433, this most likely indicates a weak cluster structure in cluster 0.

\begin{figure}[ht!]
\vskip 0.2in
\begin{center}
    \centerline{\includegraphics[width=\columnwidth]{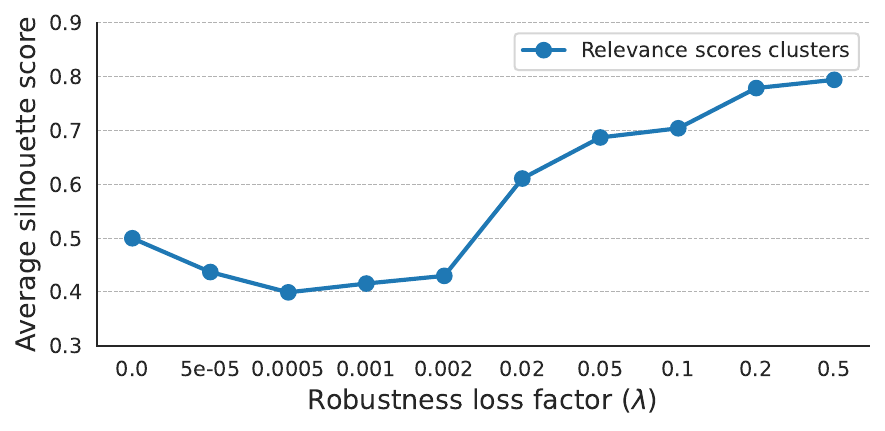}}
    \caption{Silhouette score averaged over all actions, plotted against the robustness loss factor.}
    \label{fig:silhouette}
\end{center}
\end{figure}

% 4. Figure 2, discuss the relation between the stability and clustering. Be brief
Furthermore, we tested how the quality of clustering changes under different local stability constraints, for this purpose 5 different models were trained with different robustness losses. We averaged the silhouette score over actions, for each $\lambda$. Results of this experiment are presented in Fig. \ref{fig:silhouette}. It appears that for higher $\lambda$ values the silhouette score significantly improves.

\begin{figure}[ht!]
  \vskip 0.2in
  \begin{center}
    \centerline{\includegraphics[width=\columnwidth]{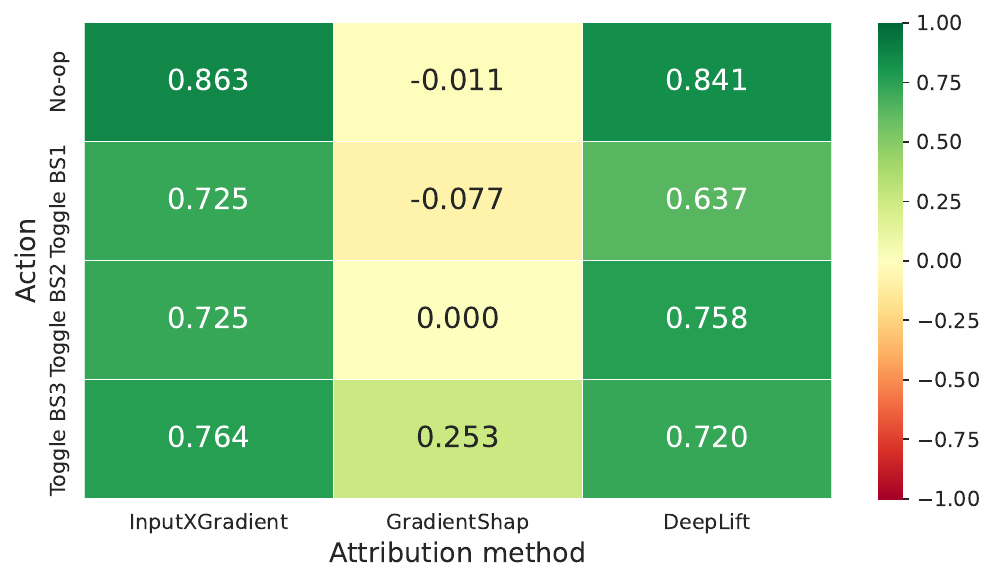}}
    \caption{Spearman correlation per action, of clustering derived attributions and other attributive \gls{xai} methods.}
    \label{fig:attribution_correlation}
  \end{center}
\end{figure}

% 5. Lastly discuss the results
Lastly, to evaluate accuracy of global explanations and validate the approach we decided to extract attributions of each concept and compare it with  well-established post-hoc methods: GradientSHAP, DeepLift and InputXGradient \cite{GradSHAP,DeepLiftIXG,inputxgradient}. We calculated the attributions by averaging sway scores over concepts, then plot Spearman correlation of obtained attributions scores with attributions from other methods, see Fig. \ref{fig:attribution_correlation}. Fig. \ref{fig:attribution_correlation}, shows that DeepLift and InputXGradient are strongly correlated with our attributions, while GradSHAP shows weak correlation at best. The key advantage of clustering over simple averaging is that it preserves the diversity of the model's behaviour, rather than collapsing all decisions into a single mean that may average out opposing values, clustering separates them into distinct typical sway profiles. Furthermore, the improvement in silhouette score with higher $\lambda$ values (Fig.~\ref{fig:attribution_correlation}) hints at relation between enforcing local stability and preserving cluster structure within the relevance scores space.

Regarding the comparison with gradient-based methods, while clustering attributions correlate with DeepLift and InputXGradient, an important distinction must be noted: relevance scores explain the final decision of the model directly, not the internal computation of the parametrizer. This makes them more immediately interpretable, as they quantify each concept's direct contribution to the output rather than propagating gradients through opaque intermediate layers, however it also introduces an element of uncertainty coming from opaqueness of the parametrizer. Lastly, in our experiments we visualized the learned bias, it provides a useful indication of which actions the policy will take more frequently; however, its effects on decontamination of the relevance scores are insignificant. In conclusion the proposed methods enable \glspl{senn} to provide both local and global intrinsic explanations, which constitute the main interpretability advantage of the approach.

\section{Conclusions}
This work is the first to apply \gls{senn}-based models to an RL problem, demonstrating that \glspl{senn} trained with \gls{ppo} achieve performance close to state-of-the-art \gls{dnn}-based methods while significantly outperforming the best heuristics. Beyond predictive performance, \glspl{senn} provide intrinsic local explanations for individual decisions, and the proposed clustering-based method successfully aggregates these into global explanations, validated through strong correlation with established gradient-based post-hoc methods. This combination of competitive performance with both local and global explanations grounded in the model's actual decision mechanism makes  \glspl{senn} a promising candidate for high-stakes domains where transparency is a prerequisite and input features are understandable.

The primary limitation is that explanation quality is sensitive to the choice of $\lambda$, which currently requires empirical tuning. For future work, one natural direction is an adaptive local stability mechanism, where the Lipschitz constraint varies with concept space density, reducing the performance cost of explainability. Additionally, further investigation into how local stability shapes the relevance score space could deepen theoretical understanding and guide deployment across broader RL settings.

\section{Impact Statement}
Our results suggest that SENNs can be applied to a broad range of RL problems, offering strong predictive performance alongside intrinsic explainability grounded in the model's internal mechanism. In practice, SENN-based RL systems could provide reliable per-timestep explanations of agent decisions and help diagnose biases in learned policies, contributing positively to safety and fairness - particularly in sensitive applications.

However, the approach carries risks that should be acknowledged. The choice of robustness loss factor strongly influences explanation quality, and no universal value guarantees consistent explanations across problems. Additionally, SENNs-based models provide per timestep decisions and do not capture the temporal aspect, which may be important for understanding long-term strategies. Lastly, our solution, in general, depends on interpretable input features or concepts that could be extracted from the input data. Lack of such could be detrimental towards understandability of provided explanations. 

Overall, SENN-based RL methods represent a meaningful step toward more transparent and trustworthy reinforcement learning, but their responsible deployment requires careful tuning and awareness.

\bibliographystyle{styles/icml2026}
\bibliography{styles/refs}

\end{document}